\pdfoutput=1
\documentclass{article} 
\usepackage{iclr2026_conference,times}

\usepackage{amsmath,amsfonts,bm}

\def\eqref#1{equation~\ref{#1}}

\def\1{\bm{1}}

\DeclareMathAlphabet{\mathsfit}{\encodingdefault}{\sfdefault}{m}{sl}
\SetMathAlphabet{\mathsfit}{bold}{\encodingdefault}{\sfdefault}{bx}{n}

\usepackage{hyperref}
\usepackage{url}
\usepackage{booktabs}
\usepackage{graphicx}
\usepackage{amsmath,amssymb}
\usepackage{algorithm}
\usepackage{algpseudocode}
\usepackage{microtype}

\title{HarnessBandit: Joint Learnability--Transferability Scheduling\\
for Multi-Harness Agentic Reinforcement Learning}

\author{\mdseries
Hongliang Wei$^{1,2,*}$ \quad
Xiaobing Tu$^{2,*,\dagger}$ \quad
Yinggui Wang$^{2}$ \quad
Zhengxi Liu$^{2}$ \\
Rongkun Xue$^{2}$ \quad
Jinkui Ren$^{2}$ \quad
Xiantao Zhang$^{2}$ \quad
Debin Zhao$^{1}$ \quad
Xiaopeng Fan$^{1,\dagger}$ \\[1.5ex]
$^{1}$ Harbin Institute of Technology \\
$^{2}$ Alibaba Cloud \\[1ex]
$^{*}$ Equal contribution. \quad
$^{\dagger}$ Corresponding authors.
}

\iclrfinalcopy

\begin{document}

\maketitle

\begin{abstract}
Language-model agents are increasingly deployed through diverse \emph{harnesses} that differ in system prompts, tool schemas, control loops, and trajectory formats. The same model can perform unevenly across these interfaces, making robustness to harness variation an important objective. A natural approach is to train a shared policy through multiple harnesses, but doing so introduces a scheduling problem: each training step should favor a harness that currently provides a useful learning signal while also producing an update that benefits the other harnesses. We develop \textbf{HarnessBandit}, an online scheduler that selects one harness per optimizer step. After a group-relative policy optimization (GRPO) update, it observes \emph{learnability}---the mean absolute advantage on the batch---and \emph{transferability}---the cosine between a low-dimensional gradient sketch of the current harness and exponential moving averages of the remaining harnesses. The two signals are fused after pooled sliding-window min--max normalization and sampled with a visit-dependent bonus and an explicit exploration floor. We train \texttt{Qwen3.5-2B} across six harnesses on ClawGym and evaluate on PinchBench (held-out tasks, in-distribution OpenClaw) and ClawEval (held-out tasks and harness). HarnessBandit improves over mixed-batch multi-harness training on both benchmarks, while training diagnostics indicate that learnability and transferability provide distinct, evolving signals. 

\end{abstract}

\section{Introduction}
\label{sec:intro}

Language-model agents interact with tools through a \emph{harness}: a combination of system prompt, tool schema, planner or loop policy, and session bookkeeping that turns a user query into a trajectory of completions and tool calls~\citep{yao2023react,yang2024sweagent,wang2025openhands}. Different harnesses induce different trajectory distributions even when the underlying task and reward function are held fixed. As a result, a policy trained exclusively through one interface need not transfer to another. Training with multiple harnesses can reduce this sensitivity, but requires deciding how their data should be allocated within and across optimizer steps.

This decision is not a standard multi-task mixture problem. The same ClawGym task can be executed under OpenClaw, Codex, Gemini CLI, or another scaffold; the \emph{task} identity is shared, while the \emph{execution channel} changes. We consider two properties of that channel. First, a batch is useful only if it produces a non-degenerate learning signal. Under group-relative policy optimization~\citep{shao2024deepseekmath,deepseekr1}, a prompt group whose sampled trajectories receive identical rewards yields advantages near zero and therefore almost no policy gradient. We call this \emph{learnability}. Second, an update computed on one harness may align with directions that other harnesses also require, or it may overfit one scaffold. We call the former \emph{cross-harness transferability}. A static allocation of training steps responds to neither quantity.

We formulate multi-harness training as a non-stationary multi-armed bandit problem, where each harness is an arm and its online utility jointly reflects learnability and cross-harness transferability. We introduce \textbf{HarnessBandit}, a scheduling layer on top of GRPO rather than a new policy objective. At each step it selects one harness for a pure rollout batch, measures how much learning signal the batch provides and how well its update aligns with the other harnesses, and combines these observations to adapt the subsequent sampling distribution while retaining exploration.

We train a \texttt{Qwen3.5-2B} policy~\citep{qwen3} with Dr.GRPO~\citep{liu2024drgrpo} on 2{,}000 ClawGym tasks~\citep{clawgym2025} crossed with six harnesses (12{,}000 prompt rows), and evaluate on PinchBench (held-out tasks under OpenClaw, an in-distribution harness) and ClawEval (held-out tasks under ClawEval's native loop, an out-of-distribution harness). HarnessBandit obtains a PinchBench mean of $0.506$, compared with $0.493$ for a two-harness OpenClaw--Codex mixture and $0.331$ for a six-harness mixture (untuned base $0.330$). On ClawEval (161 tasks, $N{=}3$), it reaches $41.6\%$ pass@3 and mean score $0.556$, compared with $38.5\%$ / $0.510$ and $29.8\%$ / $0.490$; Pass$^3$ is $26.7\%$, matching the base model and above both mixtures.

\begin{figure}[t]
    \centering
    \includegraphics[width=0.98\linewidth]{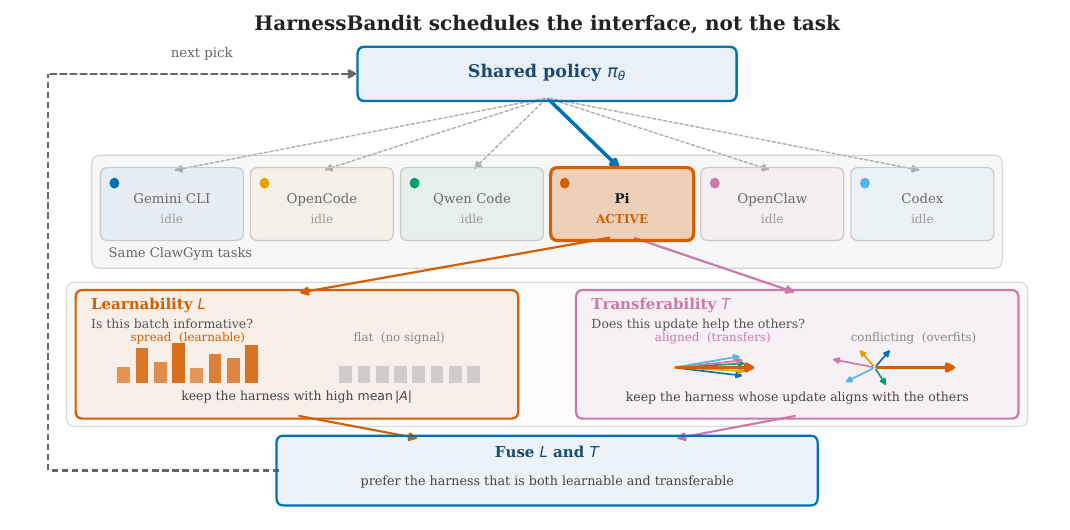}
    \caption{Principle of HarnessBandit. A shared policy sees the same tasks through six harnesses; only one harness is active per optimizer step. Learnability is the within-group reward spread (a usable mean $|A|$). Transferability is whether the current update aligns with sketches of the idle harnesses. The two scores are fused to choose the next interface. Glyphs are schematic.}
    \label{fig:overview}
\end{figure}

\paragraph{Contributions.}
\begin{enumerate}
    \item We formulate multi-harness training as a non-stationary bandit over \emph{execution interfaces}, rather than as a fixed mixture over tasks or domains.
    \item We propose two complementary trainer-native signals---local learnability and cross-harness transferability---and combine them so the scheduler prefers updates that are both informative and aligned with other interfaces.
    \item We conduct experiments on PinchBench (held-out tasks, in-distribution OpenClaw) and ClawEval (held-out tasks and harness), where HarnessBandit improves over static harness mixtures under a matched GRPO budget without collapsing to a single arm.
\end{enumerate}

\section{Related work}
\label{sec:related}

\paragraph{Agentic reinforcement learning and tool use.}
Tool-using language agents interleave reasoning with environment actions~\citep{yao2023react,schick2023toolformer,qin2024toolllm}. Subsequent work has treated the \emph{agent--computer interface} itself as a first-class design object: SWE-agent showed that scaffolding choices change success rates on software-engineering tasks~\citep{yang2024sweagent}, OpenHands and SWE-Gym provide open stacks for training and evaluating such agents~\citep{wang2025openhands,pan2024swegym}, and web and personal-assistant benchmarks stress long-horizon tool use~\citep{zhou2024webarena,yao2024taubench,liu2024agentbench}. Reinforcement learning has been applied to reasoning traces~\citep{ouyang2022instructgpt,shao2024deepseekmath,deepseekr1,yu2025dapo} and, more recently, to agent trajectories with programmatic rewards. Polar~\citep{polar2026} exposes multiple harnesses as interchangeable rollout backends behind a shared trainer. We use that interface. Our question is orthogonal to the choice of policy gradient: given several harnesses, which one should generate the next on-policy batch.

\paragraph{Multi-task learning and conflicting updates.}
When a shared model is trained on several sources, gradients can interfere~\citep{caruana1997multitask,ruder2017overview,sener2018multi,yu2020gradient}. Multi-task RL methods distill shared skills or regularize policies toward a common centroid~\citep{parisotto2016actor,teh2017distral,barreto2017successor}. Those approaches typically assume a fixed sampling distribution over tasks. In our setting the ``tasks'' are harnesses that wrap the \emph{same} underlying ClawGym instances; interference is induced by interface mismatch rather than by disjoint objectives. HarnessBandit does not project or surgically alter gradients. It chooses which harness to roll out so that the observed update is both informative and aligned with other interfaces.

\paragraph{Curriculum learning and data mixing.}
Curriculum learning sequences training material by difficulty or progress~\citep{bengio2009curriculum,graves2017automatic,narvekar2020curriculum,portelas2020curriculum}. In language-model pretraining, mixture weights over domains are often tuned offline or by a proxy model~\citep{xie2023doremi,albalak2023data}. Most such methods do not observe whether an on-policy source currently yields a non-zero advantage or whether its gradient aligns with other sources. A related two-signal bandit has been used to schedule reasoning domains~\citep{yang2026tac}. We instead allocate training steps across execution harnesses that wrap the same task collection.

\paragraph{Bandit allocation.}
Classical bandits allocate pulls under uncertainty~\citep{auer2002ucb,lattimore2020bandit}. We use an exponentially weighted value, a visit-dependent bonus, Boltzmann sampling, and a uniform floor; $c/\sqrt{n_h+1}$ is an implementation choice rather than a new regret result. The contribution is the execution-harness instantiation and its systems integration.

\paragraph{Gradient similarity and sketches.}
Cosine similarity and related kernel alignments are standard probes of whether two updates or representations point in the same direction~\citep{yosinski2014transfer,kornblith2019cka}. Storing one full gradient per harness is impractical at even 2B scale. Linear sketches, including coordinate subsampling and Rademacher projections, preserve inner products in the sense of the Johnson--Lindenstrauss lemma~\citep{johnson1984extensions,achlioptas2003database,charikar2002sketch}. HarnessBandit uses a fixed, layer-proportional subsample of the last four transformer blocks so that cosine comparisons remain well-defined across steps without materializing 190M-dimensional buffers.

\section{Problem formulation}
\label{sec:problem}

\paragraph{Tasks and harnesses.}
Let $\mathcal{X}$ be a set of agent tasks. Each task $x\in\mathcal{X}$ specifies a user query, an initial workspace, and a scalar reward $r\in[0,1]$ computed by a programmatic checker after the session ends. A \emph{harness} $h\in\mathcal{H}$ is an execution interface: it wraps $x$ with a system prompt, a tool schema, and a control loop, and it induces a trajectory distribution $\tau\sim \pi_\theta(\cdot\mid x,h)$ for a policy $\pi_\theta$. Two harnesses applied to the same $x$ can produce different tokenizations, different tool-call syntax, and different reward realizations.

Training uses a finite product $\mathcal{D}_{\mathrm{train}}=\{(x,h):x\in\mathcal{X}_{\mathrm{train}},\,h\in\mathcal{H}_{\mathrm{train}}\}$. Evaluation uses disjoint task collections $\mathcal{X}_{\mathrm{eval}}$ (PinchBench and ClawEval) that are never seen during optimization. PinchBench is executed under OpenClaw, which belongs to $\mathcal{H}_{\mathrm{train}}$; ClawEval uses a native evaluation loop that does not. The intended generalization is therefore across held-out tasks, and, for ClawEval, across interfaces as well.

\paragraph{On-policy update.}
We train with group-relative policy optimization~\citep{shao2024deepseekmath}. For a prompt $(x,h)$ the rollout stack draws $G$ trajectories, scores them, and forms advantages $A_{i,t}$ by subtracting a leave-one-out group baseline. Following Dr.GRPO~\citep{liu2024drgrpo} we do not divide by the group standard deviation, which avoids exploding updates when intra-group reward variance is near zero. The policy loss at step $t$ is the usual clipped surrogate plus a KL penalty toward a frozen reference,
\begin{equation}
  \mathcal{L}_t(\theta)
  =
  \mathbb{E}\!\left[
    -\min\!\bigl(\rho_{i,t} A_{i,t},\;
    \mathrm{clip}(\rho_{i,t},1-\epsilon,1+\epsilon)\,A_{i,t}\bigr)
  \right]
  +\beta_{\mathrm{KL}}\,\mathrm{KL}\!\bigl(\pi_\theta\,\|\,\pi_{\mathrm{ref}}\bigr),
\end{equation}
where $\rho_{i,t}$ is the importance ratio. Nothing in HarnessBandit depends on this particular surrogate; any on-policy method that yields per-token advantages and a batch gradient can supply the two observations below.

\paragraph{Scheduling objective.}
Under the bandit abstraction, the arms are $\mathcal{H}_{\mathrm{train}}$ and one arm is selected at each optimizer step. Because the shared policy changes after every update, arm utilities evolve with $\theta_t$ rather than following fixed reward distributions. A scheduler is a (possibly randomized) rule $\sigma$ that, given history $\mathcal{F}_{t-1}$, selects $h_t\in\mathcal{H}_{\mathrm{train}}$. The rollout then uses only $h_t$ for every prompt in the batch. We want $\sigma$ to trade off two instantaneous properties of $h$:
\begin{align}
  L_h(\theta)
  &=
  \mathbb{E}\!\left[\,\lvert A\rvert \;\middle|\; h,\,\theta\,\right],
  \label{eq:learnability}\\
  T_h(\theta)
  &=
  \mathbb{E}\!\left[
    \frac{1}{\lvert\mathcal{H}_{\mathrm{train}}\setminus\{h\}\rvert}
    \sum_{j\neq h}
    \cos\!\bigl(\nabla_\theta\mathcal{L}(h),\,\bar g_j\bigr)
    \;\middle|\; h,\,\theta
  \right],
  \label{eq:transferability}
\end{align}
where $\bar g_j$ is a running estimate of the gradient direction associated with harness $j$. $L_h$ is large when the group contains both successful and unsuccessful trajectories, so the policy gradient is informative. $T_h$ is large when an update on $h$ is aligned with the directions previously observed on other harnesses. A scheduler that maximizes only $L_h$ can concentrate on a harness whose advantages are large but idiosyncratic. A scheduler that maximizes only $T_h$ can prefer a harness that is well aligned yet currently uninformative ($L_h\approx 0$). HarnessBandit estimates both quantities online and selects $h_t$ to increase a convex combination of their normalized values, subject to exploration.

The quantity we ultimately care about is not $L_h$ or $T_h$ themselves, but expected reward on $\mathcal{X}_{\mathrm{eval}}$. The two signals are instruments: they are cheap to compute from the training step and are intended to allocate a fixed step budget toward updates that are both usable and shareable.

\section{Method}
\label{sec:method}

HarnessBandit wraps an unchanged GRPO trainer with an online harness scheduler. Figure~\ref{fig:overview} states the principle: the same tasks are executed through several interfaces, only one interface is active at a step, and the next draw depends on both local learnability and cross-harness transferability. Algorithm~\ref{alg:harnessbandit} gives the per-step procedure. After each update the scheduler records a local learnability signal and a sketched-gradient transferability signal, fuses them after pooled sliding-window min--max normalization, and draws the next harness. The implementation uses a proportional coordinate subsample, per-selected-step signal updates, $\tau{=}0.3$, and an explicit $\varepsilon$-floor.

\subsection{Harness-pure batches}
\label{sec:purity}

Let $\mathcal{H}_{\mathrm{train}}=\{h^{(1)},\ldots,h^{(H)}\}$. Each harness maintains an independent shuffled pool of the same ClawGym tasks, so harness identity is not confounded with task difficulty. At step $t$ the scheduler returns one name $h_t$; every prompt in the rollout batch of size $B$ is drawn from the pool of $h_t$. Purity has two consequences. First, the batch gradient is attributable to a single arm, so a cosine against other arms is well-defined. Second, leftover groups produced by oversampling filters can be stashed per harness rather than discarded, which keeps the comparison with a mixed-batch baseline honest on the number of scored trajectories.

\subsection{Learnability}
\label{sec:learnability}

After advantages are computed, each trajectory is reduced to a token-masked mean of $\lvert A_{i,t}\rvert$. Trajectory means that share a prompt group are averaged, and group means that share a harness are averaged:
\begin{equation}
  L_{h_t}
  =
  \frac{1}{\lvert \mathcal{G}_{h_t}\rvert}
  \sum_{g\in\mathcal{G}_{h_t}}
  \frac{1}{\lvert \mathcal{T}_g\rvert}
  \sum_{\tau\in\mathcal{T}_g}
  \overline{\lvert A\rvert}_\tau.
  \label{eq:L-impl}
\end{equation}
Equal weighting at the group level prevents long trajectories from dominating the signal. If every trajectory in every group of $h_t$ receives the same reward, then $L_{h_t}\approx 0$ and the step is treated as uninformative even if the absolute reward is high.

\subsection{Transferability via gradient sketches}
\label{sec:sketch}

Let $g_h=\nabla_\theta\mathcal{L}$ be the batch gradient after all microbatches of a pure $h$-step have been backwarded and before the optimizer consumes them. We retain only parameters in the last $K{=}4$ transformer blocks. Embeddings and the output head are excluded: their gradients are dominated by token frequency and therefore track which tasks were sampled rather than how the harness pushes the policy. The selected tensors still contain on the order of $10^8$ coordinates. A sketch $\hat g_h\in\mathbb{R}^d$ with $d{=}65{,}536$ is formed by a fixed random subsample whose per-tensor allocation is proportional to $\mathrm{numel}(\cdot)$. Proportional allocation keeps each tensor's contribution to $\lVert\hat g_h\rVert_2^2$ scaled by the same constant, so the sketched cosine is not biased by layer shape. Across data-parallel and context-parallel ranks the sketch is summed and renormalized; cosine similarity is scale-invariant, so the subsequent division by world size is only for interpretability.

Each harness stores an exponential moving average $\mathrm{EMA}_j$ of its unit sketch on CPU. After the current sketch is reduced,
\begin{equation}
  T_{h_t}
  =
  \frac{1}{\lvert J\rvert}
  \sum_{j\in J}
  \cos\!\bigl(\hat g_{h_t},\,\mathrm{EMA}_j\bigr),
  \qquad
  J=\{j\in\mathcal{H}_{\mathrm{train}}:j\neq h_t,\;\mathrm{EMA}_j\text{ exists}\}.
  \label{eq:T-impl}
\end{equation}
If $J$ is empty---as it is on the first visit to each arm---$T_{h_t}$ is left undefined and the score below falls back to learnability alone. The EMA is then updated with rate $\eta{=}0.3$.

\subsection{Normalization, fusion, and value update}
\label{sec:fusion}

$L_h$ and $T_h$ have different units. Each is min--max normalized over a sliding window of the last $W{=}32$ raw observations, \emph{pooled across harnesses}:
\begin{equation}
  \tilde L_h
  =
  \mathrm{clip}\!\left(\frac{L_h-\min\mathcal{W}_L}{\max\mathcal{W}_L-\min\mathcal{W}_L},\,0,1\right),
\end{equation}
and likewise for $\tilde T_h$. Per-arm windows would erase the between-arm contrasts that the bandit needs. If a window contains fewer than two values, the normalized coordinate is set to $1/2$. The instantaneous score is
\begin{equation}
  S_h
  =
  \begin{cases}
    \alpha\,\tilde L_h+(1-\alpha)\,\tilde T_h
      & \text{if both are defined,}\\[2pt]
    \tilde L_h\text{ or }\tilde T_h
      & \text{if only one is defined.}
  \end{cases}
  \label{eq:score}
\end{equation}
We use $\alpha{=}0.5$. The per-arm value is an EMA, $Q_h\leftarrow (1-\beta)Q_h+\beta S_h$ with $\beta{=}0.3$. An unvisited arm is assigned $\max_j Q_j$ at selection time so that softmax cannot starve it before the first observation.

\subsection{Sampling}
\label{sec:sampling}

The quantity that is ranked is $U_h=Q_h+c/\sqrt{n_h+1}$ with $c{=}0.2$. The sampling distribution is
\begin{equation}
  p_h
  =
  (1-\varepsilon)\,\mathrm{softmax}(U/\tau)_h
  +\varepsilon/H,
  \qquad \tau{=}0.3,\;\varepsilon{=}0.15.
  \label{eq:probs}
\end{equation}
The temperature $\tau$ controls how sharply the curriculum concentrates; the floor $\varepsilon$ guarantees a minimum rate that does not depend on $n_h$. The bonus $c/\sqrt{n_h+1}$ is the term that actually prefers under-sampled arms. Because the bonus does not grow with $t$ (unlike UCB1's $\sqrt{\ln t/n}$), an arm that stops being drawn keeps a frozen bonus and cannot rescue itself. That is why $\varepsilon$ and $c$ are not interchangeable.

The first $W_{\mathrm{up}}{=}30$ steps ignore $p$ and cycle through $\mathcal{H}_{\mathrm{train}}$ in round-robin order, so every arm has at least five observations before the bandit takes over.

\begin{algorithm}[t]
\caption{One HarnessBandit training step}
\label{alg:harnessbandit}
\footnotesize
\begin{algorithmic}[1]
\Require $t,\theta_t$, pools $\{\mathcal{D}_h\}_{h=1}^H$, values $Q$, visits $n$, sketch EMAs $E$
\If{$t<W_{\mathrm{up}}$}
    \State $h_t \gets h^{(t\bmod H)}$ \Comment{round-robin warmup}
\Else
    \State $U_h \gets Q_h+c/\sqrt{n_h+1}$; \quad
    $p_t(h) \gets (1-\varepsilon)\operatorname{softmax}(U/\tau)_h+\varepsilon/H$
    \State $h_t\sim p_t$
\EndIf
\State $\mathcal{B}_t \gets \Call{Rollout}{\mathcal{D}_{h_t},\pi_{\theta_t},B,G}$
\State compute GRPO advantages $A_t$, loss $\mathcal{L}_t$, and
$L_{h_t}\gets\Call{GroupMeanAbs}{A_t}$
\State $\widehat g_{h_t}\gets\Call{SketchAndReduce}{\nabla_\theta\mathcal{L}_t,K,d}$
\State $J\gets\{j\ne h_t:E_j\text{ is available}\}$
\If{$J\ne\varnothing$}
    \State $T_{h_t}\gets |J|^{-1}\sum_{j\in J}\cos(\widehat g_{h_t},E_j)$
\EndIf
\State normalize available $L_{h_t},T_{h_t}$; form $S_{h_t}$ by Eq.~\ref{eq:score}
\State $Q_{h_t}\gets(1-\beta)Q_{h_t}+\beta S_{h_t}$; \quad $n_{h_t}\gets n_{h_t}+1$
\State $E_{h_t}\gets\Call{UnitEMA}{E_{h_t},\widehat g_{h_t}}$
\State $\theta_{t+1}\gets\Call{OptimizerStep}{\theta_t,\nabla_\theta\mathcal{L}_t}$
\end{algorithmic}
\end{algorithm}

\subsection{Complexity}
\label{sec:complexity}

Relative to an existing GRPO step, the extra work is (i)~a gather of $d$ gradient coordinates on $K$ layers, (ii)~an all-reduce of a $d$-vector, and (iii)~$O(Hd)$ cosine evaluations. With $d{=}65{,}536$ and $H{=}6$, the EMA store occupies a few megabytes rather than several gigabytes. The policy loss, rollout, and reward evaluation are unchanged. HarnessBandit is therefore a constant-factor add-on to an existing GRPO run, not a second trainer.

\section{Experiments}
\label{sec:exp}

\subsection{Training setup}
\label{sec:setup}

\paragraph{Data.}
Training uses 2{,}000 ClawGym tasks~\citep{clawgym2025}. Each task directory contains a user query, optional workspace files, and a \texttt{reward.sh} checker that prints a scalar in $[0,1]$. We form a Cartesian product with six training harnesses---Gemini CLI, OpenCode, Qwen Code, Pi, OpenClaw, and Codex---yielding 12{,}000 prompt rows. Every method therefore sees the same task universe; harness identity is not confounded with difficulty.

\paragraph{Policy and optimizer.}
The policy is \texttt{Qwen3.5-2B}~\citep{qwen3}, trained with Polar~\citep{polar2026} and slime/Megatron GRPO~\citep{shoeybi2019megatron} on four GPUs. We use Dr.GRPO~\citep{liu2024drgrpo} (leave-one-out baseline, no standard-deviation normalization), Adam at $10^{-6}$ with a constant schedule, KL coefficient $0.05$, rollout batch size $B{=}8$, $G{=}8$ samples per prompt, and 150 optimizer steps. Dynamic sampling drops groups with zero reward standard deviation. Context parallelism is 4 and the maximum response length is 8{,}192 tokens. 

\paragraph{HarnessBandit hyperparameters.}
Unless noted: $\alpha{=}0.5$, $\beta{=}0.3$, $\tau{=}0.3$, $c{=}0.2$, $\varepsilon{=}0.15$, warmup $30$ steps, window $32$, sketch dimension $65{,}536$, last $K{=}4$ layers, sketch EMA $0.3$.

\paragraph{Comparisons.}
All trained methods share the optimizer, data product, step budget, and hardware.
\begin{itemize}
    \item \textbf{HarnessBandit}: one harness per batch, selected as in Section~\ref{sec:method}.
    \item \textbf{Mixed (6 harnesses)}: the same 12{,}000-row file with a global shuffle, so a batch of eight prompts typically mixes several harnesses. No bandit and no gradient sketch.
    \item \textbf{Mixed (OpenClaw+Codex)}: the same mixed loader restricted to two harnesses. This tests whether a smaller interface set is sufficient.
    \item \textbf{Qwen3.5-2B (base)}: the untuned checkpoint, evaluated with the same serving stack.
\end{itemize}

\subsection{Evaluation protocol}
\label{sec:eval-proto}

Held-out evaluation uses two independent benchmarks. Both use tasks disjoint from ClawGym. They differ in whether the evaluation interface belongs to $\mathcal{H}_{\mathrm{train}}$.

\paragraph{PinchBench.}
We report official-suite results on 147 tasks, one run per task, in the official OpenClaw Docker environment~\citep{pinchbench2026}. OpenClaw is one of the six training harnesses, so this suite is out-of-distribution in tasks and in-distribution in interface. The metric is the mean automated task score in $[0,1]$.

\paragraph{ClawEval.}
We follow the official protocol on the \texttt{general} split (161 tasks) with $N{=}3$ independent trials~\citep{claweval2026}. Execution uses ClawEval's own think--act--observe loop and per-task tool schemas, not any member of $\mathcal{H}_{\mathrm{train}}$. The suite is therefore out-of-distribution in both tasks and interface. Pass$^3$ is the fraction of tasks that pass on all three trials; pass@3 is the fraction that pass on at least one trial; the average score is the mean of the 161 per-task scores.

\subsection{Results}
\label{sec:main-results}

\begin{table}[t]
\caption{Held-out evaluation after ClawGym training. PinchBench: official 147-task OpenClaw suite (held-out tasks, in-distribution harness). ClawEval: \texttt{general} split, 161 tasks, native loop (held-out tasks and harness), $N{=}3$. Pass$^3$: all three trials pass. Pass@3: at least one trial passes. Avg score: mean of 161 per-task scores. Same 150-step GRPO budget for trained methods.}
\label{tab:main}
\centering
\small
\setlength{\tabcolsep}{3.5pt}
\begin{tabular}{lcccc}
\toprule
& PinchBench & \multicolumn{3}{c}{ClawEval} \\
\cmidrule(lr){2-2} \cmidrule(lr){3-5}
Method & Mean & Pass$^3$ & Pass@3 & Avg score \\
\midrule
Qwen3.5-2B (base) & $0.330$ & $\mathbf{43/161}$ ($26.7\%$) & $58/161$ ($36.0\%$) & $0.534$ \\
Mixed (OC+Codex) & $0.493$ & $32/161$ ($19.9\%$) & $48/161$ ($29.8\%$) & $0.490$ \\
Mixed (6 harnesses) & $0.331$ & $35/161$ ($21.7\%$) & $62/161$ ($38.5\%$) & $0.510$ \\
HarnessBandit & $\mathbf{0.506}$ & $\mathbf{43/161}$ ($26.7\%$) & $\mathbf{67/161}$ ($41.6\%$) & $\mathbf{0.556}$ \\
\bottomrule
\end{tabular}
\end{table}

Table~\ref{tab:main} summarizes the two benchmarks. On PinchBench (held-out tasks, OpenClaw), HarnessBandit records $0.506$, a $+1.3$ point edge over the two-harness OpenClaw--Codex mixture ($0.493$) and a $+17.6$ point gap over both the six-harness mixture ($0.331$) and the untuned base model ($0.330$). The six-harness mixed-batch run is essentially tied with the base model, so shuffling all six interfaces into mixed batches does not improve the official mean; restricting the mixed loader to OpenClaw and Codex accounts for nearly all of the trained-policy gain on this suite.

On ClawEval (held-out tasks, native loop) the ranking depends on the metric. Pass$^3$ ties HarnessBandit with the untuned base model at $43/161$ ($26.7\%$) and places both mixed schedules below the base model ($35$ and $32$ tasks). Pass@3 and mean score separate the checkpoints: HarnessBandit reaches $67/161$ ($41.6\%$) and $0.556$, against $62/161$ ($38.5\%$) / $0.510$ for the six-harness mixture, $58/161$ ($36.0\%$) / $0.534$ for the base model, and $48/161$ ($29.8\%$) / $0.490$ for the two-harness mixture. Thus both static mixtures underperform the base model on Pass$^3$ and on mean score; only HarnessBandit matches the base model on the strict official criterion and exceeds it on pass@3 and average score. PinchBench category means (Appendix~\ref{app:pinch}) are highest for HarnessBandit on analysis, log analysis, productivity, research, skills, writing, and integrations. The two-harness mixture is higher on coding, CSV analysis, meeting analysis, and memory. The six-harness mixture wins no category and scores zero on writing.

\subsection{Bandit behavior}
\label{sec:bandit-analysis}

Figure~\ref{fig:bandit} (Appendix~\ref{app:bandit}) shows the induced schedule. After warmup the mass is non-uniform: OpenClaw, Pi, and OpenCode end near $p=0.20,0.20,0.18$, whereas Qwen Code and Codex end near $0.13,0.14$. Final visit counts range from $21$ to $32$, a mild reallocation rather than collapse.

\begin{figure}[t]
    \centering
    \includegraphics[width=\linewidth]{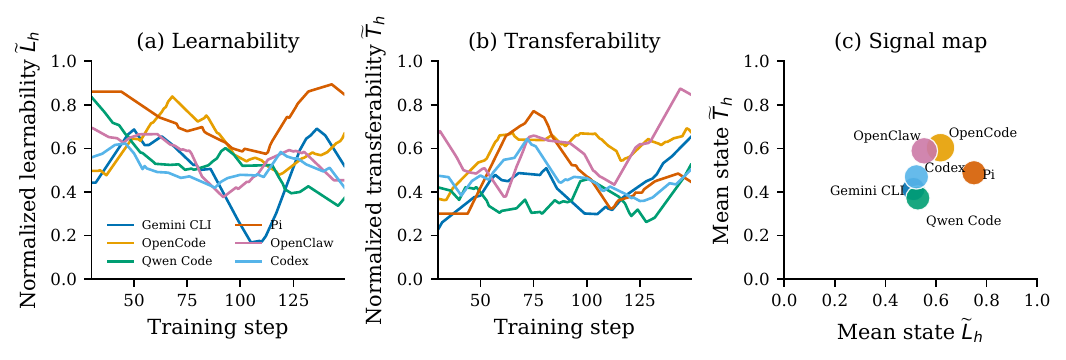}
    \caption{Learnability--transferability analysis for HarnessBandit after warmup. (a,b) 31-step moving averages of the pooled-window normalized signals. (c) Each harness's mean step-wise scheduler state; marker area is proportional to mean sampling probability.}
    \label{fig:signal-analysis}
\end{figure}

Figure~\ref{fig:signal-analysis} shows that the two signals give different rankings. Pi has the highest mean normalized learnability ($0.75$) but only mid-range transferability ($0.49$); OpenCode is high on both ($0.62,0.60$); OpenClaw's transferability ($0.59$) exceeds its learnability ($0.55$); Qwen Code is lower on transferability ($0.37$) than learnability ($0.53$). Thus transferability is not a monotone transformation of learnability in this run. Marker area reflects mean sampling probability $p_h$, which is largest for OpenCode ($0.21$) and OpenClaw ($0.18$).

Appendix~\ref{app:cosines} plots directed pairwise gradient cosines. They mostly fluctuate around zero, with both positive and negative intervals and no uniformly beneficial source harness. This fine-grained geometry explains why the scalar transferability state changes over time.

\section{Conclusion}
\label{sec:conclusion}

Multi-harness agentic RL requires a decision that GRPO does not make: which interface should generate the next on-policy batch. We argued that the decision should depend on learnability of the current update and transferability of that update across harnesses. HarnessBandit estimates both from GRPO advantages and sketched last-layer gradients, fuses them after shared-window normalization, and samples a harness-pure batch with a UCB-softmax rule.

After 150 GRPO steps on a ClawGym $\times$ six-harness product, a \texttt{Qwen3.5-2B} policy trained with HarnessBandit records a higher official PinchBench mean than both mixed schedules (held-out tasks, in-distribution OpenClaw). On ClawEval (held-out tasks and harness) it matches the untuned base model on Pass$^3$ ($26.7\%$) while improving pass@3 and mean score, and it exceeds both mixtures on all three ClawEval metrics. The induced schedule is a mild reweighting, not a collapse to one harness.

\bibliography{refs}
\bibliographystyle{iclr2026_conference}

\appendix
\counterwithin{figure}{section}
\counterwithin{table}{section}
\section{Hyperparameters}
\label{app:hparams}

Table~\ref{tab:hparams} lists the GRPO and scheduler settings shared by every trained method. The optimizer, batch size, sample count, step budget, and hardware are matched so that Table~\ref{tab:main} isolates harness allocation. The bandit knobs ($\alpha,\beta,\tau,c,\varepsilon$, warmup, window, sketch) are the values used for the reported HarnessBandit run; mixed-batch baselines ignore them.

\begin{table}[h]
\caption{Training hyperparameters shared by all GRPO arms.}
\label{tab:hparams}
\centering
\small
\begin{tabular}{ll}
\toprule
Item & Value \\
\midrule
Base model & Qwen3.5-2B \\
Algorithm & GRPO / Dr.GRPO (no std.\ normalization) \\
Adam learning rate & $1{\times}10^{-6}$, constant \\
KL coefficient / type & $0.05$ / low-var KL \\
Rollout batch size $B$ & 8 \\
Samples per prompt $G$ & 8 \\
Optimizer steps & 150 \\
Max response / prompt tokens & 8{,}192 / 40{,}000 \\
Context parallel size & 4 \\
GPUs & 4 \\
Dynamic sampling & drop zero-std groups; oversample batch 8 \\
\midrule
$\alpha$, $\beta$, $\tau$, $c$, $\varepsilon$ & $0.5$, $0.3$, $0.3$, $0.2$, $0.15$ \\
Warmup / window & 30 / 32 \\
Sketch dim / layers / EMA & $65{,}536$ / last 4 / $0.3$ \\
Sketch mode & proportional coordinate subsample \\
Training harnesses & Gemini CLI, OpenCode, Qwen Code, Pi, OpenClaw, Codex \\
\bottomrule
\end{tabular}
\end{table}

\section{PinchBench category means}
\label{app:pinch}

Table~\ref{tab:pinch-cat} breaks the official PinchBench mean into the suite's task categories. Cell sizes are identical across methods, so differences are not an artifact of a different task mix. HarnessBandit is highest on analysis, integrations, log analysis, productivity, research, skills, and writing. The two-harness OpenClaw--Codex mixture is higher on coding, CSV analysis, meeting analysis, and memory---categories that align with those two interfaces. The six-harness mixture wins no category; its writing cell is exactly zero, which we treat as a recorded score rather than evidence that the mixture cannot write. The overall row reproduces Table~\ref{tab:main}.

\begin{table}[h]
\caption{Official PinchBench mean score by category. Cell counts are identical across methods (e.g., coding $n{=}14$, CSV $n{=}26$). Overall means match Table~\ref{tab:main}.}
\label{tab:pinch-cat}
\centering
\scriptsize
\setlength{\tabcolsep}{3.5pt}
\begin{tabular}{lcccc}
\toprule
Category & Base & Mixed (OC+Codex) & Mixed (6) & HarnessBandit \\
\midrule
analysis & $0.368$ & $0.529$ & $0.366$ & $\mathbf{0.579}$ \\
coding & $0.305$ & $\mathbf{0.673}$ & $0.558$ & $0.661$ \\
csv\_analysis & $0.148$ & $\mathbf{0.433}$ & $0.241$ & $0.352$ \\
integrations & $0.135$ & $0.191$ & $0.299$ & $\mathbf{0.364}$ \\
log\_analysis & $0.397$ & $0.546$ & $0.328$ & $\mathbf{0.568}$ \\
meeting\_analysis & $0.389$ & $\mathbf{0.457}$ & $0.300$ & $0.444$ \\
memory & $0.370$ & $\mathbf{0.913}$ & $0.750$ & $0.637$ \\
productivity & $0.336$ & $0.348$ & $0.427$ & $\mathbf{0.485}$ \\
research & $0.418$ & $0.452$ & $0.229$ & $\mathbf{0.477}$ \\
skills & $0.229$ & $0.367$ & $0.560$ & $\mathbf{0.676}$ \\
writing & $0.483$ & $0.581$ & $0.000$ & $\mathbf{0.596}$ \\
\midrule
overall & $0.330$ & $0.493$ & $0.331$ & $\mathbf{0.506}$ \\
\bottomrule
\end{tabular}
\end{table}

\section{Bandit schedule}
\label{app:bandit}

\begin{figure}[h]
    \centering
    \includegraphics[width=\linewidth]{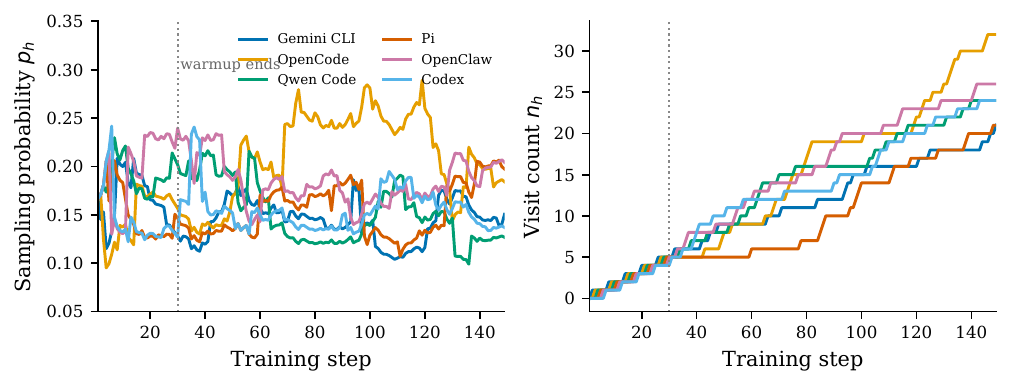}
    \caption{HarnessBandit on the reported run. Left: sampling probabilities. Right: visit counts. The vertical line marks the end of the 30-step round-robin warmup.}
    \label{fig:bandit}
\end{figure}

Figure~\ref{fig:bandit} shows the sampling distribution $p_h$ and the visit counts $n_h$ on the reported run. The first 30 steps are a round-robin warmup, so every harness is observed before softmax takes over. After that line the mass is non-uniform but stays away from a one-arm collapse: OpenClaw, Pi, and OpenCode receive more probability, while Qwen Code and Codex remain above the $\varepsilon$-floor. Visit counts at the end of training span 21--32 (Appendix~\ref{app:state}), which is a mild reallocation of a 150-step budget.

\section{Raw online signals}
\label{app:signals}

\begin{figure}[h]
    \centering
    \includegraphics[width=\linewidth]{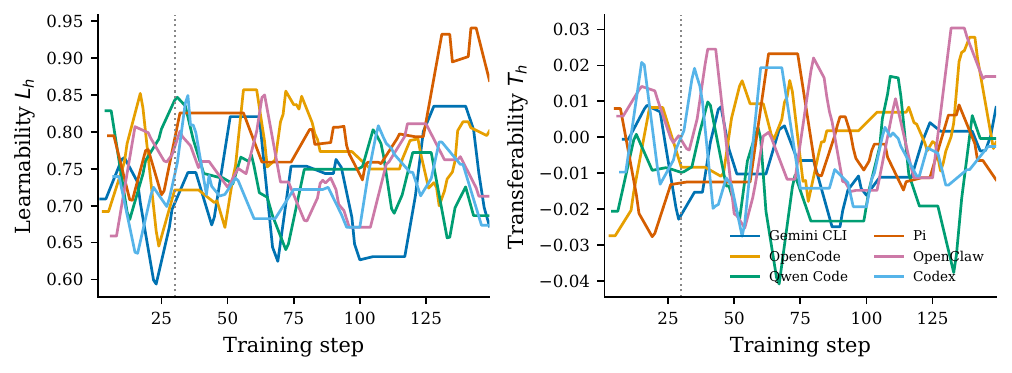}
    \caption{Online signals on the HarnessBandit run (7-step moving average). Left: learnability $L_h$. Right: transferability $T_h$.}
    \label{fig:signals}
\end{figure}

Figure~\ref{fig:signals} plots the un-normalized observations that later become $\tilde L_h$ and $\tilde T_h$. Learnability $L_h$ is a mean absolute advantage and stays clearly positive; transferability $T_h$ is an average cosine and fluctuates in a narrow band around zero. The two traces therefore live in different units, which is why Section~\ref{sec:fusion} applies a shared sliding-window min--max before fusion. Rankings also differ: a harness can look learnable on the left panel while sitting near the bottom of the right panel. Figure~\ref{fig:signal-analysis} in the main text shows the same run after normalization.

\section{Pairwise gradient geometry}
\label{app:cosines}

\begin{figure}[h]
    \centering
    \includegraphics[width=\linewidth]{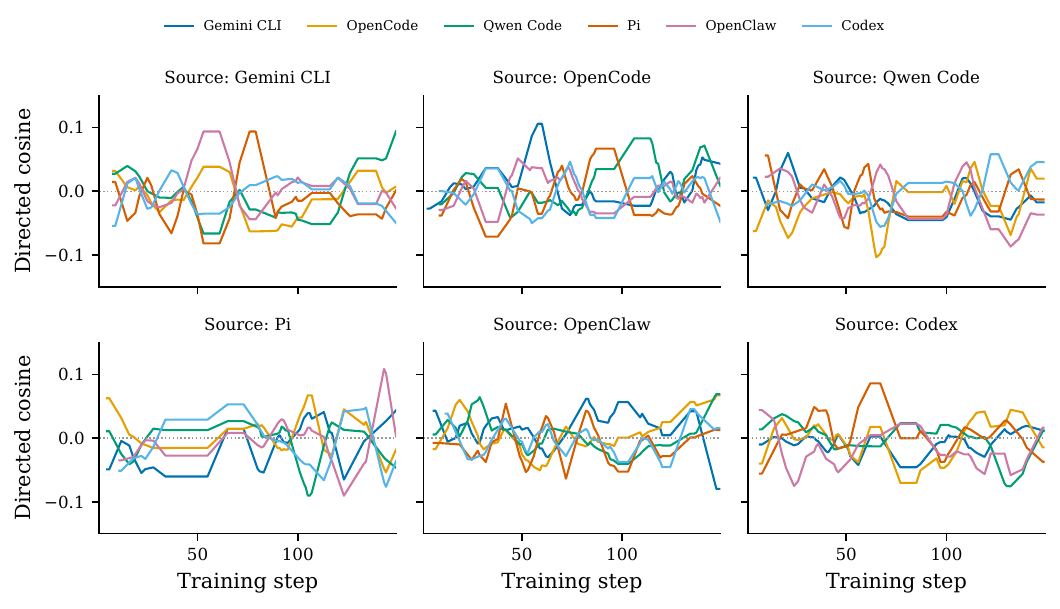}
    \caption{Pairwise analysis of directed current-to-peer gradient cosines (10-step moving average). Each panel fixes the source harness; line color denotes the peer EMA.}
    \label{fig:pairwise-cosines}
\end{figure}

The scalar $T_{h_t}$ averages the current sketch against every other harness EMA. Figure~\ref{fig:pairwise-cosines} unrolls that average into directed pairs. Each panel fixes the source harness; each line is the cosine from that source's current sketch to one peer EMA. The traces move through both positive and negative intervals and do not identify a source that is aligned with all peers at all times. That geometry is why the fused transferability state in Figure~\ref{fig:signal-analysis} evolves rather than locking onto a single ``helpful'' interface.

\section{Final HarnessBandit state}
\label{app:state}

These are the terminal $p_h$ and $n_h$ of Figure~\ref{fig:bandit}, recorded after the last optimizer step. Sampling probabilities are OpenClaw $0.204$, Pi $0.197$, OpenCode $0.184$, Gemini CLI $0.151$, Codex $0.137$, Qwen Code $0.127$. Visit counts are OpenCode $32$, OpenClaw $26$, Qwen Code $24$, Codex $24$, Gemini CLI $21$, Pi $21$ (148 observations after warmup). Uniform allocation over six arms would give $p_h=1/6\approx 0.167$ and about 25 visits each; the observed schedule is a tilt toward OpenClaw, Pi, and OpenCode, not a collapse to one arm. Pi's visit count is low relative to its final $p_h$ because probability mass arrived late.

\end{document}